%% file: aaai2026.tex
\documentclass[letterpaper]{article} % DO NOT CHANGE THIS
\usepackage{aaai2026}  % DO NOT CHANGE THIS
\usepackage{times}  % DO NOT CHANGE THIS
\usepackage{helvet}  % DO NOT CHANGE THIS
\usepackage{courier}  % DO NOT CHANGE THIS
\usepackage[hyphens]{url}  % DO NOT CHANGE THIS
\usepackage{graphicx} % DO NOT CHANGE THIS
\usepackage{enumitem}
\usepackage{booktabs}
\usepackage{multirow}
\usepackage{multicol}
\usepackage{makecell}
\usepackage{comment}
\usepackage{subcaption}
\usepackage{listings}
\usepackage{amsmath}
\usepackage{utfsym}
\usepackage[utf8]{inputenc} % allow utf-8 input
\usepackage{amsfonts}       % blackboard math symbols
\usepackage{nicefrac}       % compact symbols for 1/2, etc.
\usepackage{microtype}      % microtypography
\usepackage{titletoc}
\usepackage{natbib}  % DO NOT CHANGE THIS AND DO NOT ADD ANY OPTIONS TO IT
\usepackage{caption} % DO NOT CHANGE THIS AND DO NOT ADD ANY OPTIONS TO IT
\usepackage{algorithm}
\usepackage{algorithmic}

\usepackage{newfloat}
\usepackage{listings}
\DeclareCaptionStyle{ruled}{labelfont=normalfont,labelsep=colon,strut=off} % DO NOT CHANGE THIS
\floatstyle{ruled}
\newfloat{listing}{tb}{lst}{}
\floatname{listing}{Listing}
\title{Affordance-R1: Reinforcement Learning for Generalizable Affordance Reasoning in Multimodal Large Language Model}
\author {
    Hanqing Wang\textsuperscript{\rm 1,8,}\equalcontrib,
    Shaoyang Wang\textsuperscript{\rm 2,}\equalcontrib,
    Yiming Zhong\textsuperscript{\rm 3},
    Zemin Yang\textsuperscript{\rm 3},
    Jiamin Wang\textsuperscript{\rm 3}
    Zhiqing Cui\textsuperscript{\rm 5},
    Jiahao Yuan\textsuperscript{\rm 4},
    Yifan Han\textsuperscript{\rm 7},
    Mingyu Liu\textsuperscript{\rm 6,8},
    Yuexin Ma\textsuperscript{\rm 3,}\thanks{Corresponding Author.}
}
\affiliations {
    \textsuperscript{\rm 1}The Hong Kong University of Science and Technology (GZ),
    \textsuperscript{\rm 2}National University of Singapore,
    \textsuperscript{\rm 3}ShanghaiTech University\\
    \textsuperscript{\rm 4} East China Normal University,
    \textsuperscript{\rm 5}  Nanjing University of Information Science \& Technology,
     \textsuperscript{\rm 6}Zhejiang University\\
    \textsuperscript{\rm 7}Institute of Automation, Chinese Academy of Science,
    \textsuperscript{\rm 8}Shanghai AI Laboratory

    hwang201@connect.hkust-gz.edu.cn
}

\usepackage{bibentry}
\begin{document}

\maketitle

\input{sec/abstract}

\input{sec/Introduction}
\input{sec/RelatedWork}

\input{sec/Dataset}

\input{sec/method}

\input{sec/Experiment}

\input{sec/conclusion}

\bibliography{aaai2026}
\input{sec/sup}
\end{document}

%% file: sec/abstract.tex
\begin{abstract}
% OpenAI o1 and DeepSeek R1 achieve or even surpass human expert-leve performance in complex domains like mathematics and science, with reinforcement learning (RL) and reasoning playing a crucial role.The core of R1 lies in its rule-based reward formulation, which leverages tasks with deterministic ground-truth answers to enable precise and stable reward computation

% Recently, DeepSeek R1 has shown that reinforcement learning (RL) can substantially improve the reasoning capabilities of Large Language Models (LLMs) through a simple
% yet effective design. The core of R1 lies in its rule-based
% reward formulation, which leverages tasks with deterministic ground-truth answers to enable precise and stable reward computation

% Traditional methods for affordance grounding rely on supervised fine-tuning with categorical labels and simple descriptions, limiting its out-of-domain generalization and lacking explicit reasoning processes.

Affordance grounding focuses on predicting the specific regions of objects that are associated with the actions to be performed by robots. It plays a vital role in the fields of human-robot interaction, human-object interaction, embodied manipulation, and embodied perception. Existing models often neglect the affordance shared among different objects because they lack the Chain-of-Thought(CoT) reasoning abilities, limiting their out-of-domain (OOD) generalization and explicit reasoning capabilities. To address these challenges, we propose Affordance-R1, the first unified affordance grounding framework that integrates cognitive CoT guided Group Relative Policy Optimization (GRPO) within a reinforcement learning paradigm. Specifically, we designed a sophisticated affordance function, which contains format, perception, and cognition rewards to effectively guide optimization directions. Furthermore, we constructed a high-quality affordance-centric reasoning dataset, ReasonAff, to support training. Trained exclusively via reinforcement learning with GRPO and without explicit reasoning data, Affordance-R1 achieves robust zero-shot generalization and exhibits emergent test-time reasoning capabilities. Comprehensive experiments demonstrate that our model outperforms well-established methods and exhibits open-world generalization. To the best of our knowledge, Affordance-R1 is the first to integrate GRPO-based RL with reasoning into affordance reasoning. The code of our method and our dataset is released on \textit{https://github.com/hq-King/Affordance-R1}.

% Affordance grounding, as a fundamental concept, focuses on predicting the specific regions of objects that are associated with the actions to be performed by robots. It plays a vital role in the fields of human-robot interaction (HRI), human-object interaction (HOI), embodied manipulation and embodied perception. Existing methods often overlook the common knowledge within different objects because they lack reasoning abilities and world knowledge, limiting their out-of-domain (OOD) generalization and explicit reasoning capabilities. To address these challenges, we propose \textbf{Affordance-R1}, the first unified affordance grounding framework that integrates cognitive chain-of-thought (CoT)-guided Group Relative Policy Optimization (GRPO) within a reinforcement learning (RL) paradigm. Specifically, we first construct a high-quality affordance-centric grounding dataset \textbf{ReasonAff}. We then design a sophisticated affordance reward to effectively guide optimization directions. Trained exclusively via reinforcement learning with GRPO and without explicit reasoning data, \textbf{Affordance-R1} achieves robust zero-shot generalization and exhibits emergent test-time reasoning capabilities. Comprehensive experiments demonstrate that our model outperforms well-established methods and exhibits open-world generalization with affordance reasoning. To the best of our knowledge, \textbf{Affordance-R1} is the first to integrate GRPO-based RL with reasoning into affordance grounding. The code of our method and our dataset will be publicly released soon.

\end{abstract}

%% file: sec/Introduction.tex
\section{Introduction}

\begin{figure}[t]
    \centering
    \includegraphics[width=0.9\linewidth]{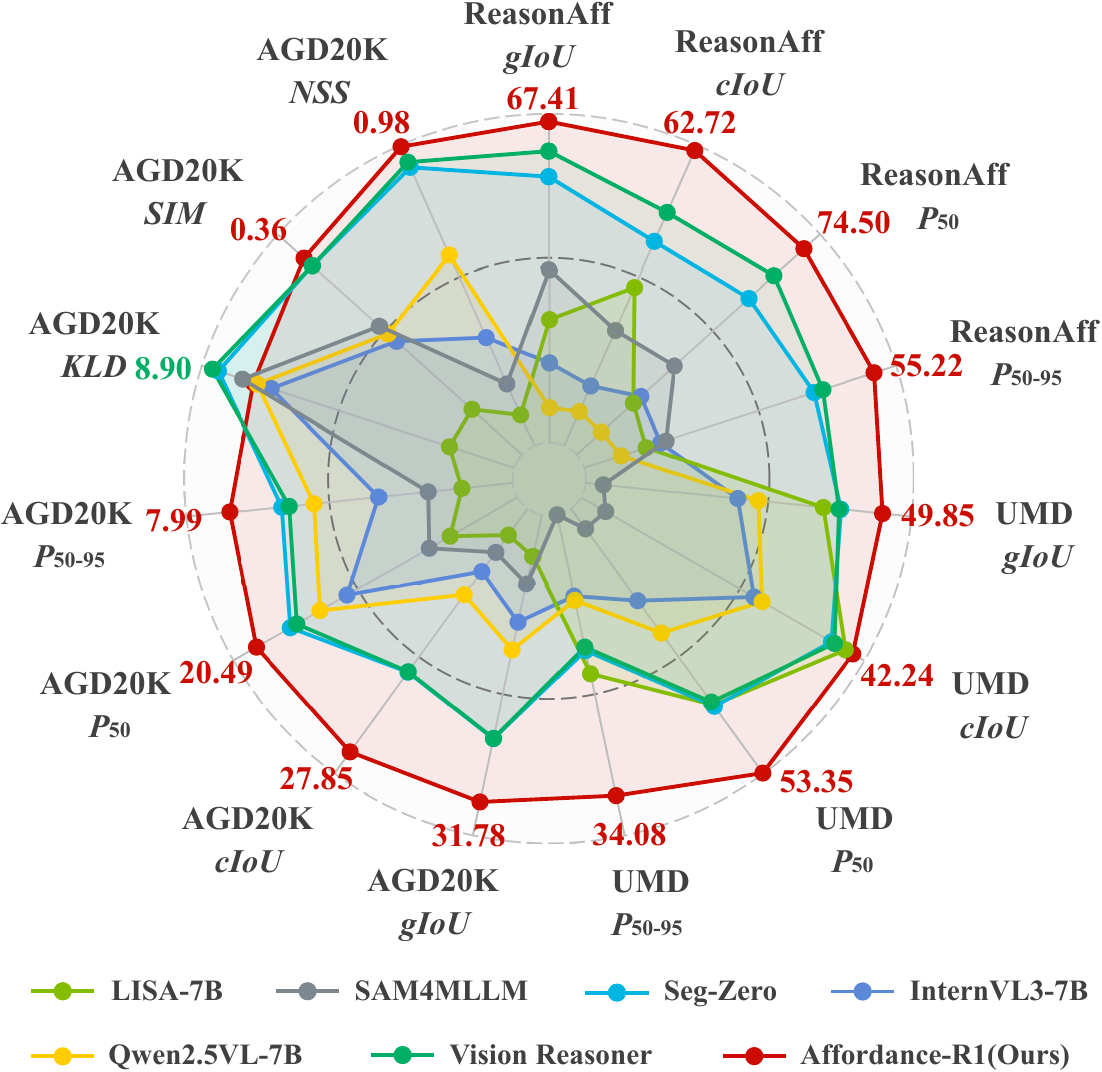}
    \caption{Affordance-R1 demonstrates extraordinary affordance reasoning ability and powerful generalization ability.}
    \label{teaser}
\end{figure}
% affordance 及 affordance grounding的概念和重要意义
Affordance is a crucial lens through which humans and embodied agents interact with various objects of the physical world, reflecting the possibility of where and how to act. Given an open-ended, complex, and implicit task instruction specified in natural language, affordance grounding aims to highlight the actionable possibilities of these objects, linking visual perception with robotic manipulation. 

%当前affordance grounding的常规技术和局限及挑战
Recent efforts have made remarkable progress in affordance learning, such as extracting affordance knowledge from human-object-interaction (HOI) images~\cite{yang2023grounding,wang2025dag,yang2024egochoir,luo2024visual,wang2025grace, Rai_2024_CVPR}, human videos~\cite{ma2025glover++,luo2023learning,chen2023affordance}, and 3D perception modeling approaches such as object and scene point clouds~\cite{deng20213d,chu20253d,nguyen2023open,Delitzas_2024_CVPR} and 3D Gaussian Splatting~\cite{3DAffordSplat}. However, these methods cannot actively reason about complex and implicit user intentions. Real-world physical interactions often require models to understand the human intention and reason about: \textit{``What object can afford this? Why can this object afford such an affordance? Where is the affordance area?''}. Specifically, given a kitchen scene and the question ``How would you reheat the food?'', the model must reason deeply to identify that the oven can heat food and requires the ``openable'' affordance. This lack of affordance reasoning creates a gap in real-world applications. Some research~\cite{yu2025seqafford,qian2024affordancellm} has utilized MLLM reasoning abilities to assist affordance grounding, but they only provide final affordance areas without the reasoning process—they cannot explain why an object affords such capabilities. To address this limitation, reinforcement learning offers a promising solution by enabling step-by-step reasoning through reward feedback, helping models understand both the answer and the reasoning process. Recent advances~\cite{openaio1,guo2025deepseekr1,Liu2025SegZeroRG,Shen2025VLMR1AS,Liu2025VisionReasonerUV,Huang2025VisionR1IR} have demonstrated this capability through verifiable reward mechanisms. However, these models focus primarily on object-level reasoning and cannot handle embodied perception tasks requiring fine-grained analysis, such as affordance reasoning.
%围绕我们的算法是如何解决上述局限的，给出我们方法的技术亮点，这里注意一定是要把我们解决了什么核心问题及为什么我们的技术设计可以解决该问题解释清楚，不要只说我们怎么做的。

To fill this gap, we propose \textbf{Affordance-R1}, a reinforcement learning framework that enhances affordance grounding models with deep reasoning capabilities. We employ GRPO to fine-tune MLLMs without supervised training, investigating their self-evolution potential to develop reasoning abilities rather than relying on explicitly annotated processes. To closely link reasoning with affordance grounding, we design rewards from cognitive and perceptual perspectives:  perception rewards and affordance recognition rewards. Inspired by \textit{``Think twice before you act''}, we add a rethinking reward to help the model verify its reasoning process, addressing the transparency issue in current affordance models. Additionally, a box-num reward ensures the model outputs all possible affordance areas. Through these integrated rewards, Affordance-R1 achieves comprehensive reasoning at both perceptual and cognitive levels.
%为了支持我们的目标，目前数据集的局限，然后突出我们另一个数据集的贡献，这里要把我们数据的概况和亮点介绍清楚。

% However, current datasets fall short in supporting complex affordance reasoning. They tend to be overly straightforward and simplistic, lacking the subtleties required for implicit inference. Moreover, they often fail to integrate real-world situational contexts and inadvertently reveal the affordance elements. Additionally, these datasets are specifically designed for training visual segmentation models and cannot be seamlessly integrated into the instruction-tuning training of MLLMs. To better enhance the MLLM's capability for affordance grounding and improve generalization, we construct a high-quality dataset \textit{\textbf{ReasonAff}}, which consists of fine-grained affordance masks and reasoning-based implicit instructions, which can inspire models to think. And it is general enough to be used for MLLM-based instruction-tuning.
To facilitate such reasoning capabilities, existing datasets are insufficient for complex affordance reasoning. They are overly simplistic, lack real-world contextual complexity, and are specifically tailored for training visual segmentation models, making them unsuitable for MLLM instruction tuning.To address these limitations, we construct ReasonAff, a high-quality dataset with fine-grained affordance masks and reasoning-based implicit instructions that promote deep affordance understanding, specifically tailored for MLLM training. We utilize GPT-4o~\cite{gpt4} to construct the implicit instructions by providing it with an HOI image related to the affordance and the original instruction to help the agent better understand \textit{``affordance''} and alleviate hallucination problems.
%我们整个贡献的总结：模型贡献，技术细节贡献，数据集贡献，performance贡献
% \textbf{Affordance-R1} demonstrates exceptional performance on both in-domain and out-of-domain data, which is crucial for real-world deployment. Furthermore, \textbf{Affordance-R1} maintains robust visual QA capabilities without the need for VQA training data. Experimental results show that \textbf{Affordance-R1} exhibits strong test-time reasoning capabilities and achieves superior generalization performance compared to models of the same scale. To summarize, our contributions are as follows:

Through the synergy of our reinforcement learning framework and reasoning-oriented dataset, \textbf{Affordance-R1} demonstrates exceptional performance on both in-domain and out-of-domain data, which is crucial for real-world deployment. Furthermore, \textbf{Affordance-R1} maintains robust visual QA capabilities without the need for VQA training data. Experimental results show that \textbf{Affordance-R1} exhibits strong test-time reasoning capabilities and achieves superior generalization performance compared to models of the same scale. To summarize, our contributions are as follows:

\begin{itemize}
    \item We introduce \textbf{Affordance-R1}, which is capable of generating explicit reasoning alongside the final answer. With the help of proposed affordance reasoning reward, which contains \textit{format}, \textit{perception}, and \textit{affordance recognition} reward, it achieves robust zero-shot generalization and exhibits emergent test-time reasoning capabilities.
    
    \item We construct a high-quality affordance dataset \textbf{ReasonAff} for MLLM-based instruction-tuning, which is crucial for embodied perception and reasoning.
    
    % \item \textbf{Sophisticated Affordance Reasoning Reward Function:} We designed a sophisticated affordance reasoning reward function, which contains \textit{format}, \textit{perception}, and \textit{affordance recognition} reward, to effectively guide the model to learn in a unified Reinforcement Learning (RL) paradigm.

    \item We implement extensive experiments to demonstrate the effectiveness of our learning pipeline and observe noticeable gains over baselines with strong generalization capability, which highlights the effectiveness and adaptability of our approach in real-world applications.
\end{itemize}

%% file: sec/RelatedWork.tex
\section{Related Work}

\subsection{Affordance Learning} 
The concept of affordance was popularized by psychologist James Gibson~\cite{gibson1977theory}, which reveals how embodied agents should interact with objects in dynamic, complex, and physical environments. Many researchers have made great efforts in affordance learning. Specifically, some works utilize affordance to link perception with robotic manipulations~\cite{tang2025uad,tong2024oval,ma2025glover++,ju2024robo,wu2025ragnet} and grasping~\cite{wei2025afforddexgrasp,zhang2023affordance}.
Other studies, from a perceptual perspective, focus on endowing robots with an understanding of the affordance of objects and have explored numerous methods to obtain affordance knowledge from demonstrations, such as learning from HOI images~\cite{yang2023grounding,gao2024learning,shao2024great}, human videos~\cite{ma2025glover++}, and 3D perception modeling approaches including object~\cite{deng20213d,qian2024affordancellm,yu2025seqafford,chu20253d,nguyen2023open} and scene~\cite{Delitzas_2024_CVPR} point clouds and 3DGS~\cite{3DAffordSplat}. With the remarkable progress of LLMs, impressive reasoning capabilities have been demonstrated that can simulate human thinking. Some studies have explored how to transfer the inherent reasoning ability of LLMs to affordance learning. These works~\cite{qian2024affordancellm,yu2025seqafford,chu20253d} adopt the strategy of introducing a special token into the vocabulary of LLMs and then utilize the embedding of this special token to perform affordance grounding. However, they still fail in generalization and cannot perform well when encountering OOD data, because they only establish a mapping between the affordance areas and the special token and cannot grasp general affordance knowledge. To address this issue, we utilize the GRPO~\cite{shao2024deepseekmath} algorithm to conduct a post-training process on the multimodal large language model, enabling the model to think and reason like humans to perform affordance perception.
 \begin{figure*}[ht]
    \centering
    \includegraphics[width=0.8\linewidth]{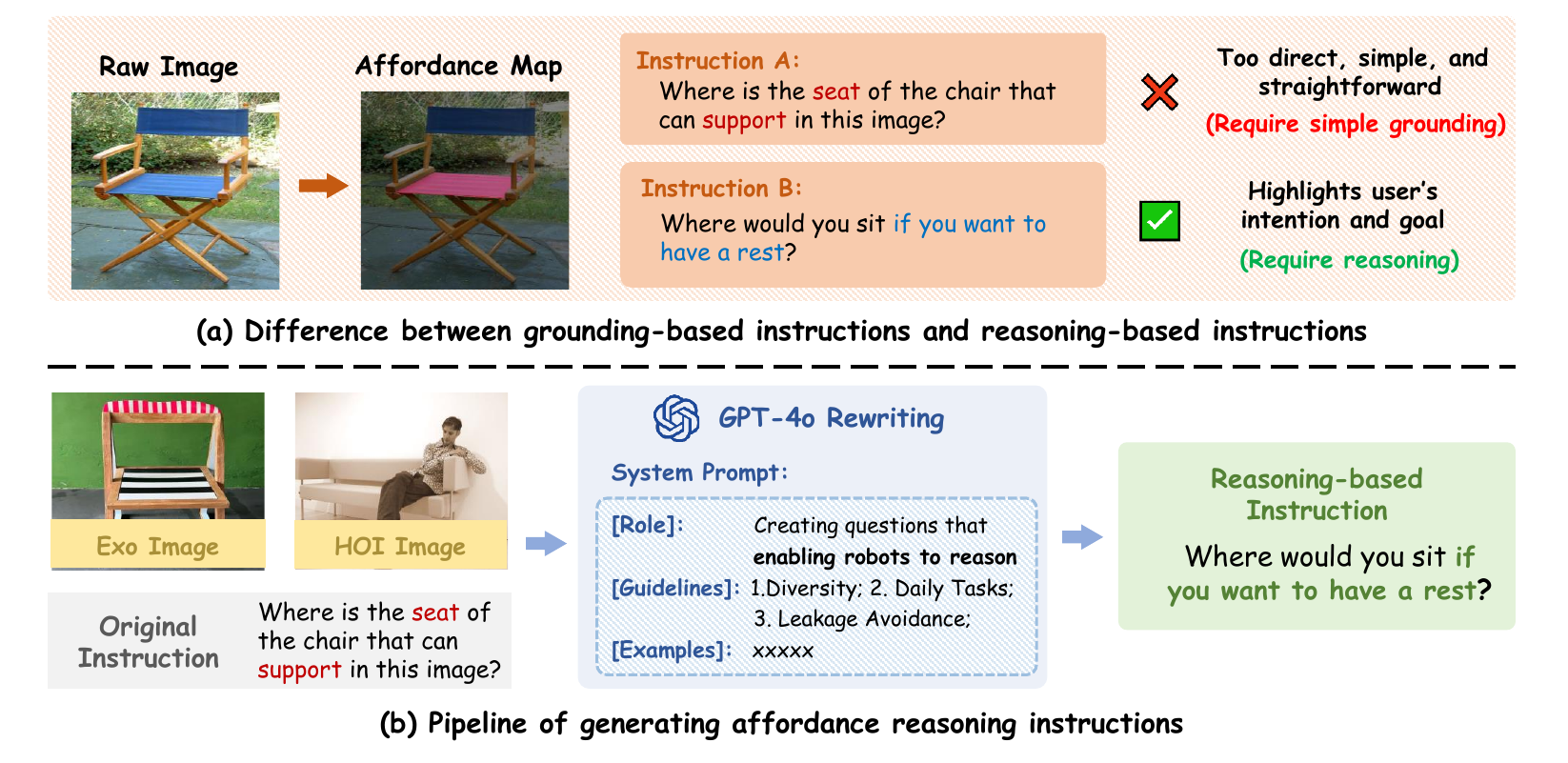}
    \caption{Affordance reasoning instruction generation and comparison. (a) Comparison between grounding-based and reasoning-based instructions. Instruction A directly asks for the faucet handle location (simple grounding), while Instruction B asks how to interact with the faucet to achieve opening (requires reasoning). (b) Pipeline for generating affordance reasoning instructions using GPT-4o to rewrite original instructions based on exo images, HOI images, and system prompts with guidelines for diversity, daily tasks, and leakage avoidance. The used prompt and statistical information of ReasonAff can be seen in our Appendix.}
    \label{datasetpipline}
\end{figure*}
\subsection{Multimodal Large Language Models} 
MLLMs~\cite{yang2025qwen3,gpt4} have made remarkable progress, which can achieve human-like or even superhuman intelligence in many aspects, such as visual understanding, generation, and multimodal reasoning. However, for many practical applications, such as segmentation and grounding, these models lack the necessary fine-grained perception required for detailed visual tasks. To address this issue, research efforts~\cite{wang2024visionllm,lan2024text4seg,wu2024visionllm} enable the localization of specific regions within images by encoding spatial coordinates as tokens, improving the models' ability to reason about precise areas within the visual data. Moreover, OpenAI o1~\cite{openaio1} introduces inference-time scaling by extending the Chain-of-Thought (CoT) reasoning process, significantly enhancing its multimodal reasoning performance. DeepSeek-R1~\cite{guo2025deepseekr1} further utilizes the GRPO~\cite{shao2024deepseekmath} algorithm to advance the reasoning ability, achieving superior performance with only a few thousand RL training steps. Several recent works~\cite{Shen2025VLMR1AS,Liu2025SegZeroRG,Huang2025VisionR1IR,Liu2025VisionReasonerUV,song2025maniplvm,ouyang2025motion,zhou2025r1,pan2025medvlm,zhang2025tinyllava,feng2025video} have expanded this success into fine-grained visual tasks. However, these works primarily address high-level object reasoning and do not consider fine-grained part-level, especially affordance-level understanding.

Addressing this limitation, this paper aims to endow MLLMs with general affordance-aware perception by enabling them to interpret and interact with objects through reasoning in context-sensitive scenarios.

% Despite these advancements, the majority of existing MLLMs are primarily focused on scene-level and object-level understanding, lacking the ability to recognize and reason about fine-grained affordances in diverse semantic contexts. Recent research~\cite{qian2024affordancellm,yu2025seqafford,chu20253d} has explored adding a special token $'<SEG>'$ or $'<AFF>'$ to the vocabulary of MLLMs to achieve affordance perception capability, but this design necessitates extensive data to fine-tune both the MLLM and the segmentation decoder, and may even compromise the pixel precision of the original segmentation models and destroy the original model's dialogue ability, causing catastrophic forgetting and thereby resulting in poor generalization. 

\begin{table}[t]
\centering
\setlength{\tabcolsep}{1.36mm}{
       \resizebox{\linewidth}{!}{
\begin{tabular}{@{}l|ccccc@{}}
\toprule
 Dataset    & \#Object  & \#Aff  & \#Diversity &  \#Reasoning   &  \#Q\&A  \\
\midrule
UMD  & 17  & 7 & $\usym{2613}$ &$\usym{2613}$  &$\usym{2613}$  \\
IIT-AFF  &  10 & 9 & $\usym{2613}$ & $\usym{2613}$   &$\usym{2613}$  \\
ADE-Af    & 150  & 7  & $\usym{2613}$& $\usym{2613}$  &$\usym{2613}$  \\
PAD  & 72 & 31   & $\usym{2613}$ & $\usym{2613}$   &$\usym{2613}$ \\
PADv2   & 103 & 39  & $\usym{2613}$ & $\usym{2613}$     &$\usym{2613}$   \\
AGD20K &50  & 36   &$\usym{2613}$ &$\usym{2613}$   &$\usym{2613}$ \\
InstructPart & 48  &30   & $\usym{2613}$ & $\usym{2613}$   &$\usym{2613}$ \\
Ours & 48  &30    & $\checkmark$ & $\checkmark$ & $\checkmark$ \\
\bottomrule
\end{tabular}}}

\caption{Comparison of Existing 2D Affordance Dateset with Ours. $\#Diversity$: diverse contextual instructions. $\#Obj$: number of object categories. $\#Aff$: number of affordance categories. $\#Q\&A$: Q\&A instruction-tuning for MLLM. 
}
\label{com_data}

\end{table}
% \subsection{Reasoning Models.} 
% Recently, Large Language Models (LLMs) have exhibited remarkable reasoning capabilities. By extending the length of the Chain-of-Thought (CoT) reasoning process, OpenAI o1~\cite{openaio1} introduces inference-time scaling, significantly enhancing its reasoning performance. In the research community, several studies have attempted to achieve test-time scaling through various approaches, including process-based reward models, reinforcement learning (RL), and search algorithms. Notably, the recent DeepSeek-R1~\cite{guo2025deepseekr1}, which utilizes the GRPO~\cite{shao2024deepseekmath} algorithm, achieves superior performance with only a few thousand RL training steps. Building on advancements in the LLM community, several recent works~\cite{Shen2025VLMR1AS,Liu2025SegZeroRG,Huang2025VisionR1IR,Liu2025VisionReasonerUV,song2025maniplvm,ouyang2025motion,zhou2025r1,pan2025medvlm,zhang2025tinyllava,feng2025video} have attempted to leverage the reasoning capabilities of MLLMs. However, these works primarily address high-level object reasoning and do not consider fine-grained part-level, especially affordance-level understanding. To fill this gap, our Affordance-R1 is designed to enhance affordance-level reasoning through pure reinforcement learning without Supervised Fine-Tuning (SFT).

%% file: sec/Dataset.tex
\begin{figure*}[t]
    \centering
    \includegraphics[width=\linewidth]{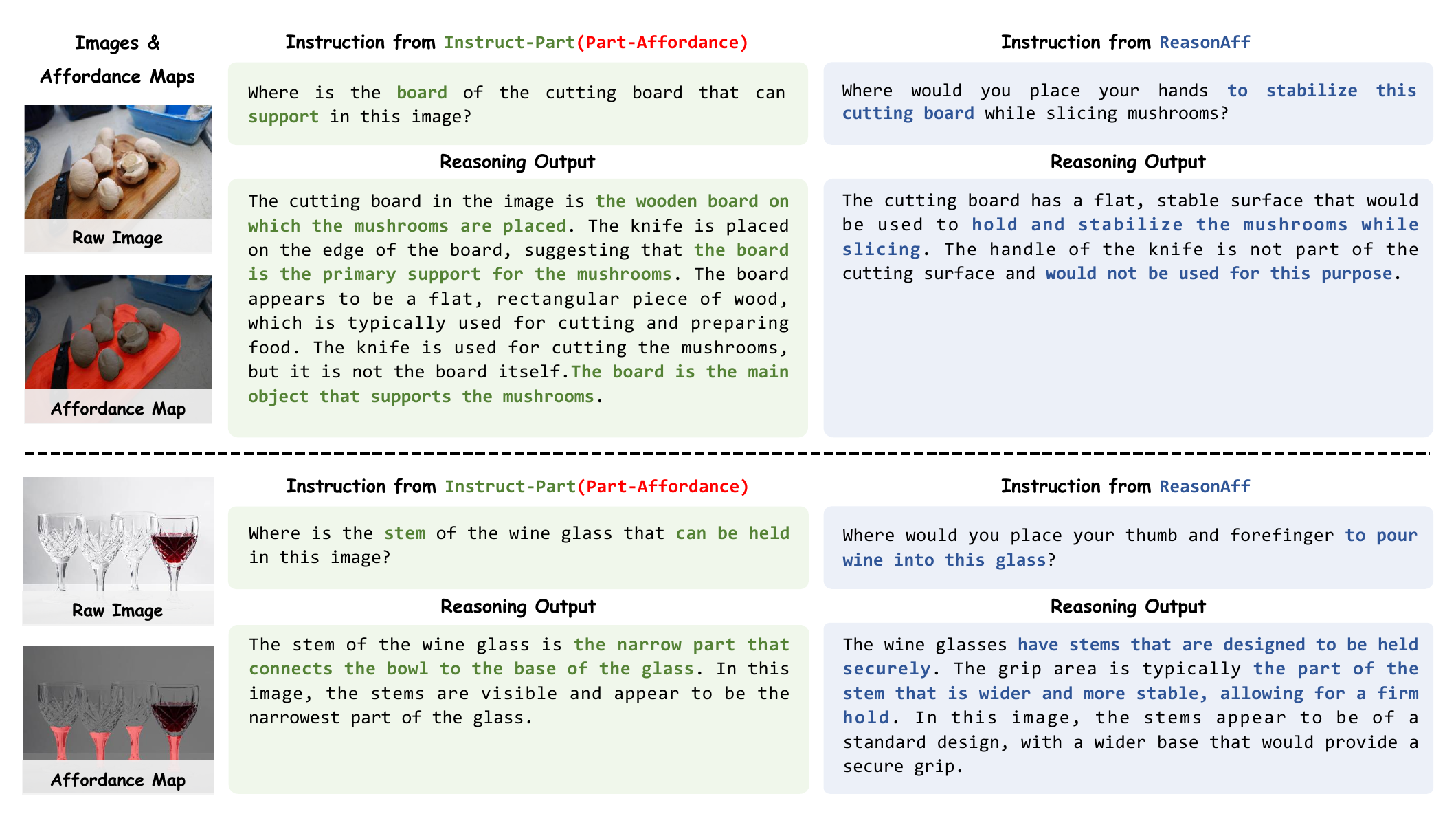}
    \caption{Comparison of instructions and reasoning outputs between ReasonAff and Instruct-Part datasets on the same images.}
    \label{data}
\end{figure*}
  
\section{Dataset}
\label{reasonaff}
Previous affordance-centric datasets fall short in supporting complex affordance reasoning. Moreover, these datasets are specifically designed for training visual segmentation models (e.g., SAM~\cite{ravi2024sam}), making them difficult to seamlessly integrate into the instruction fine-tuning of multimodal large language models (MLLMs). As a result, models trained on such datasets tend to rely on grounding rather than in-depth reasoning. This prevents them from acquiring generalizable affordance knowledge, severely undermining their generalization capabilities.

\begin{table}[h]
  \centering
    \resizebox{\linewidth}{!}{
      \begin{tabular}{c|ccc|cccc}
        \toprule
        \multirow{2}{*}{Source}& \multicolumn{3}{c|}{\textbf{Results on AGD20K }} &\multicolumn{4}{c}{\textbf{Results on UMD }} \\
        & KLD$\downarrow$ & SIM$\uparrow$ & NSS$\uparrow$ &gIoU$\uparrow$  & cIoU$\uparrow$ & $P_{50-95}$$\uparrow$ & $P_{50}$$\uparrow$  \\
        \midrule
        Instruct-Part &10.79 & 0.30 & 0.89&44.37& 38.06 & 26.24 & 47.13 \\
        ReasonAff  &\textbf{9.73} & \textbf{0.36}& \textbf{0.98}&\textbf{49.85}& \textbf{42.24}& \textbf{34.08}& \textbf{53.35} \\
        \bottomrule
       \end{tabular}}
  \caption{Evaluating Cross-Dataset Generalization for Affordance reasoning.}
\label{Crossdata_dataset}
\end{table}
% \begin{table}[h]
%   \centering
%     \resizebox{\linewidth}{!}{
%       \begin{tabular}{c|cccc|cccc}
%         \toprule
%         \multirow{2}{*}{Source}& \multicolumn{4}{c|}{\textbf{Results on AGD20K }} &\multicolumn{4}{c}{\textbf{Results on UMD }} \\
%         & KLD$\downarrow$ & SIM$\uparrow$ & NSS$\uparrow$ &gIoU  & cIoU & $P_{50}$ & $P_{50-95}$  \\
%         \midrule
%         Instruct-Part &10.79 & 0.30 & 0.89&44.37& 38.06 & 26.24 & 47.13 \\
%         ReasonAff  &9.73 & 0.36& 0.98&49.85& 42.24& 34.08& 53.35 \\
%         \bottomrule
%        \end{tabular}}
%   \caption{Evaluating Cross-Dataset Generalization for Affordance reasoning.}
% \label{Crossdata_dataset}
% \end{table}
To better enhance the affordance grounding ability of MLLMs and improve their generalization performance, we have constructed the high-quality dataset ReasonAff, which can be utilized for MLLM instruction tuning. Specifically, we construct ReasonAff based on Instruct-Part~\cite{wan2025instructpart}. As shown in Figure~\ref{datasetpipline} (b), we rewrite the instructions in the Instruct-Part dataset because we find the instructions are too direct and simple, and there are many sentences with \textbf{consistent structures} and many sentences are completely \textbf{identical}, which may limit the reasoning ability of the model. We utilize GPT-4o~\cite{gpt4} to rewrite the instructions by providing it with an HOI image related to the affordance and the original instruction to alleviate hallucination issues and avoid identical instructions to enhance \textbf{diversity}. Specifically, for a given binary mask of affordance, we determine its bounding box $(x1,y1,x2,y2)$ by extracting the leftmost, topmost, rightmost, and bottommost pixel coordinates. Additionally, we compute the centroid of the mask as point coordinates $(x_p,y_p)$. We show the comparison of ReasonAff with previous datasets in Table~\ref{com_data}, and more dataset details are provided in the Appendix.

As can be seen in Figure~\ref{data}, we present the different reasoning output (highlight areas) between the original Instruct-Part Affordance-related instructions and our reasoning-based instructions. Our implicit instructions based on reasoning can better enhance the reasoning ability of the model compared to previous instructions, enabling the model to learn more general affordance knowledge through the reasoning process and improve its generalization ability, as demonstrated by our experimental results shown in Table~\ref{Crossdata_dataset}. The model trained on the reasoning-based \textbf{ReasonAff} dataset shows better performance and generalization on OOD datasets.

% \begin{figure}
%     \centering
%     \includegraphics[width=0.9\linewidth]{images/sentence_length_violin_plot.pdf}
%     \caption{Sentence Length Distribution.}
%     \label{sentence}
% \end{figure}

%% file: sec/method.tex
 \begin{figure*}[t]
    \centering
    \includegraphics[width=0.9\linewidth]{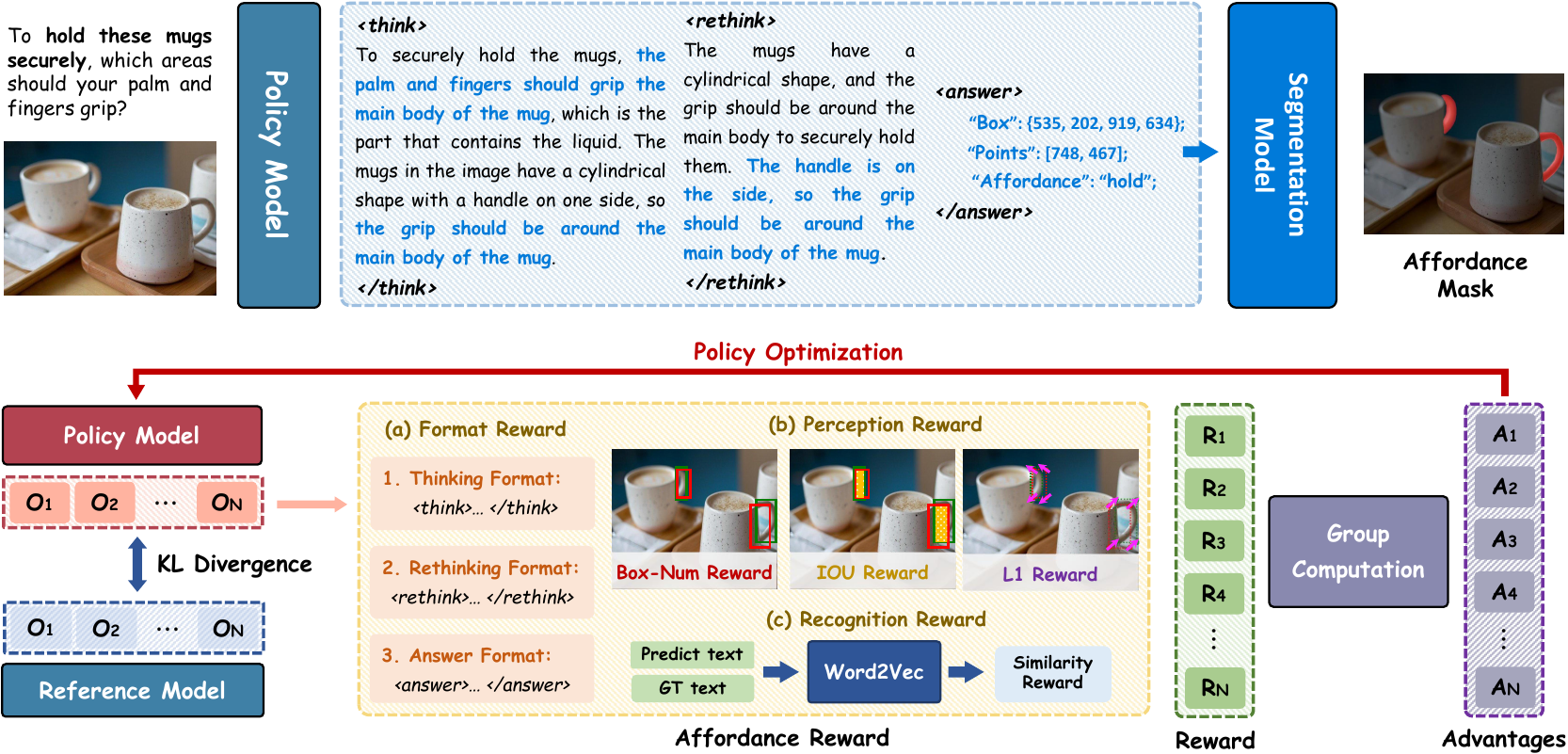}
    \caption{\textbf{Affordance-R1 framework overview}. The model processes queries through policy-based reasoning with $<think>$ and $<rethink>$ stages to generate affordance predictions. The policy optimization uses a sophisticated reward system comprising (a) format rewards for reasoning structure, (b) perception rewards for spatial accuracy (Box-Num, IOU, L1), and (c) recognition rewards for semantic similarity, enabling effective GRPO-based training for affordance reasoning.}
    \label{Pipeline}
\end{figure*}
\section{Affordance-R1 Framework}

\subsection{Overview}
We provide an overview of our proposed method \textbf{Affordance-R1}. The task we address is a reasoning-based visual affordance grounding problem, where the model is tasked with localizing functional areas on objects based on implicit and complex instructions. Formally, given a textual instruction T and a target image I, the model $\mathcal{F}$ is expected to output the affordance area $\mathcal{A}{ff}$, defined as $\mathcal{A}{ff} = \mathcal{F}(T, I)$.
Our method consists of two stages as shown in Figure~\ref{Pipeline}. In the first stage, we directly employ rule-based reinforcement learning GRPO~\cite{shao2024deepseekmath} without SFT to enhance the model's inherent reasoning abilities. Additionally, we introduce a carefully designed affordance reward containing format, perception, and recognition components, to encourage the model to think and rethink about the image before providing final answers. In the second stage, we extract the output bounding boxes and points from \textbf{Affordance-R1}, which are then used as prompts for state-of-the-art segmentation models to produce fine-grained affordance masks.

\subsection{Architecture}
Following Seg-Zero~\cite{Liu2025SegZeroRG}, \textbf{Affordance-R1} adopts a two-stage strategy comprising a reasoning model and a segmentation model. The overall architecture is illustrated in Figure~\ref{Pipeline}. Specifically, given an image $\mathbf{I}$ and a high-level text instruction $\mathbf{T}$, \textbf{Affordance-R1} $\mathcal{F}$ generates an interpretable reasoning process and subsequently produces the expected output corresponding to $\mathbf{T}$. The model output is represented in a structured format, from which we extract the bounding boxes $\mathbf{B}$ and points $\mathbf{P}$ to serve as input to segmentation models such as SAM~\cite{kirillov2023segment}. This process can be formulated as follows:
\begin{equation}
     (\{\mathbf{B}_i, \mathbf{P}_i\})_{i=1}^{N} = \mathcal{F}(\mathbf{I}, \mathbf{T}).
\end{equation}
Subsequently, the affordance masks ${A}_{ff}$ are predicted by the segmentation model $\mathcal{M}$ using the extracted bounding boxes $\mathbf{B}$ and points $\mathbf{P}$:
\begin{equation}
     \mathbf{A}_i = \mathcal{M}(\mathbf{B}_i, \mathbf{P}_i).
\end{equation}

\subsection{Group Relative Policy Optimization (GRPO)}
Unlike reinforcement learning algorithms such as PPO~\cite{schulman2017proximal}, which require an additional critic model to estimate policy performance, GRPO~\cite{shao2024deepseekmath} directly compares groups of candidate responses, thereby eliminating the need for a separate critic network. Given a question $q$, GRPO~\cite{shao2024deepseekmath} samples $N$ candidate responses $\{o_1, o_2, \ldots, o_N\}$ from the policy $\pi_\theta$ and evaluates each response $o_i$ using a reward function $R(q, o_i)$, which quantifies the quality of the candidate response in the context of the given question. To determine the relative quality of these responses, GRPO~\cite{shao2024deepseekmath} normalizes the rewards by computing their mean and standard deviation, and subsequently derives the advantage as:
\begin{equation}
    A_i = \frac{r_i - \text{mean}\{r_1, r_2, \ldots, r_N\}}{\text{std}\{r_1, r_2, \ldots, r_N\}},
\end{equation}
where $A_i$ represents the advantage of candidate response $o_i$ relative to other sampled responses within the group.GRPO~\cite{shao2024deepseekmath} encourages the model to generate responses with higher advantages by optimizing the policy $\pi_\theta$ through the following objective:
\begin{align}
    \mathcal{J}_{GRPO}&(\theta) = \mathbb{E}[{\{o_i\}_{i=1}^N\sim\pi_{\theta_{old}}(q)}] \\
    &\frac{\sum_{i=1}^N\left\{\min[s_1 A_i,\ s_2  A_i]-\beta\mathbb{D}_{KL}[\pi_\theta||\pi_{ref}]\right\}}{N} \\
    s_1 & = \frac{\pi_\theta(o_i|q)}{\pi_{\theta_{old}}(o_i|q)}; \quad s_2  = \text{clip}\left(s_1,1+\epsilon,1-\epsilon\right).
\end{align}

\subsubsection{Reward Functions.} As can be seen in Figure~\ref{Pipeline}, we designed a sophisticated affordance reward system that contains \textit{format, perception, and recognition} rewards to better guide the optimization of affordance reasoning.

\textbf{Format Reward.} We utilize the format reward to ensure the model's response strictly adheres to the required format. It can be divided into three parts: \textbf{1) Thinking Reward:} To force the model to think deeply before answering, we add the format \textit{$<think>$ Thinking Process Here $</think>$} to constrain the model; \textbf{2) Rethinking Reward:} Inspired by the proverb: \textit{``Think twice before you act''}, we add the rethinking reward \textit{$<rethink>$ Rethinking Process Here $</rethink>$} to force the model to evaluate the thinking process itself, which double-checks the correctness of the reasoning process; \textbf{3) Answer Reward:} \textit{$<answer>$ Final Answer Here $</answer>$}. 

\textbf{Perception Reward.} To help the model ground the affordance area, we utilize the perception reward, which mainly contains: \textbf{1) IoU Reward:} We calculate the Intersection over Union (IoU) between output bounding boxes and ground truth bounding boxes. If $\text{IoU} > 0.5$, the reward is 1; otherwise, the reward is 0; \textbf{2) L1 Reward:} We compute the L1 distance between output and ground truth bounding boxes (including points). If the $\text{L1 distance} < 10$, the reward is 1; otherwise, the reward is 0; \textbf{3) Box-Num Reward:} We introduce the box-num reward to ensure the model outputs all possible affordance areas.

\textbf{Affordance Recognition Reward.} As the ancient wisdom states, \textit{``to know what it is and to know why it is''}, affordance reasoning requires not only perception but also recognition. Specifically, we use the word2vec model to calculate affordance text similarity. If $\text{similarity} > 0.8$, the reward is 1; otherwise, the reward is 0.

%% file: sec/Experiment.tex
% \begin{figure}[t]
%     \centering
%     \includegraphics[width=\linewidth]{images/sentence_length_violin_plot.pdf}
%     \caption{\textbf{Statistics of ReasonAff.}Here we show the sentence length distribution of ReasonAff and . }
%     \label{data}
% \end{figure}
% \begin{table*}[t]
%   \centering
%   \scalebox{1.0}{
%       \begin{tabular}{c|cccc}
%         \toprule
%         \multirow{2}{*}{Source}& \multicolumn{4}{c}{\textbf{Results on AGD20K Dataset}} \\

%         & KLD$\downarrow$ & SIM$\uparrow$ & NSS$\uparrow$\\
%         \midrule
%         Instruct-Part &10.79 & 0.30 & 0.89& \\
%         ReasonAff  &9.73 & 0.36& 0.98& \\
%         \midrule
%         \multirow{2}{*}{Source}& \multicolumn{4}{c}{\textbf{Results on UMD Dataset}} \\
%         &gIoU  & cIoU & $P_{50}$ & $P_{50-95}$  \\
%         \midrule
%         Instruct-Part &44.37& 38.06 & 26.24 & 47.13 \\
%         ReasonAff  &49.85& 42.24& 34.08& 53.35\\
%         \bottomrule
%       \end{tabular}}
%   \caption{Evaluating Cross-Dataset Generalization for Affordance reasoning.}
% \label{Crossdata}
% \end{table*}

\begin{table}[t]
  \centering
    \resizebox{\linewidth}{!}{
      \begin{tabular}{l|cccccc}
        \toprule
         Model & LLM &Reasoning &gIoU$\uparrow$  & cIoU$\uparrow$ & $P_{50}$ $\uparrow$ & $P_{50-95}$$\uparrow$  \\
        \midrule
        VLPart & \usym{2613} & \usym{2613} & 4.21& 3.88& 1.31 & 0.85 \\
        OVSeg & \usym{2613} & \usym{2613} &16.52  & 10.59 & 9.89 & 4.12  \\
        SAN & \usym{2613} & \usym{2613} &10.21 & 13.45 & 7.18 & 3.17  \\ 
        LISA-7B & \checkmark & \usym{2613}&38.17 & 40.58 & 33.62 & 19.69 \\ 
        SAM4MLLM & \checkmark & \usym{2613} &45.51 & 33.64 & 43.48 & 22.79  \\ 
        AffordanceLLM & \checkmark & \usym{2613} & 48.49  & 38.61 & 42.11 & 20.19 \\
        InternVL3-8B & \checkmark & \usym{2613} & 31.79 & 24.68 & 35.41 & 21.93  \\

        Qwen2.5VL-7B & \checkmark & \usym{2613} & 25.18 & 20.54 & 26.00 & 15.82  \\ 
        Seg-Zero & \checkmark & \checkmark &59.26 & 48.03 & 61.33 & 45.87  \\ 
        Vision Reasoner & \checkmark & \checkmark &63.04 & 52.70 & 67.33 & 47.23  \\ 
        \textbf{Affordance-R1}(Ours) & \checkmark & \checkmark &\textbf{67.41}& \textbf{62.72} & \textbf{74.50} & \textbf{55.22} \\ 
        \bottomrule
      \end{tabular}
  }

  \caption{Affordance reasoning comparison on ReasonAff.}
\label{main_result}
\end{table}
\section{Experiment}
This section provides a comprehensive evaluation of our proposed framework, \textbf{Affordance-R1}. We first describe the experimental settings, including datasets, baseline methods, evaluation metrics, and implementation details. Next, we present the quantitative analysis of the experimental results. Additionally, we conduct ablation studies to demonstrate the effectiveness of each component of our method.

\subsection{Experimental Settings}
\subsubsection{Dataset and Out-of-Domain Datasets.} As mentioned in Section~\ref{reasonaff}, we construct a high-quality dataset \textbf{ReasonAff} based on the Instruct-Part~\cite{wan2025instructpart} dataset. We train our model on this dataset, and to assess our model's generalization capability, we conduct experiments to evaluate its performance under OOD scenarios. Specifically, we leverage subsets from the UMD Part Affordance dataset~\cite{UAD} and AGD20K~\cite{luo2022learning} as our OOD benchmarks for affordance task evaluation. 
\begin{figure}[t]
    \centering
    \includegraphics[width=\linewidth]{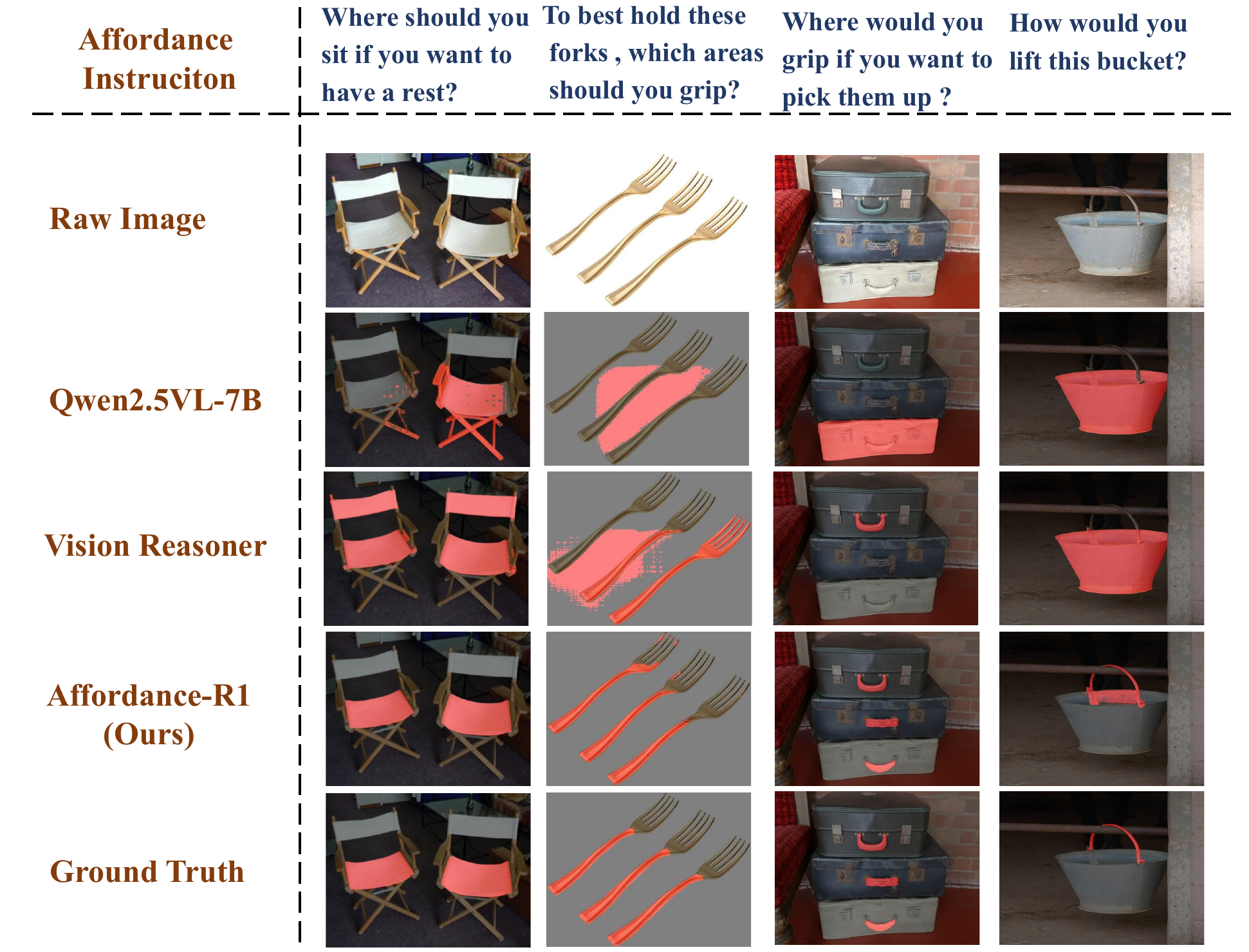}
    \caption{Qualitative Comparison of Affordance Reasoning  }
    \label{vis_reasonaff}
\end{figure}
For the UMD Part Affordance dataset~\cite{UAD}, to better assess the zero-shot performance of different models, we select all objects from all categories. Since one in every three frames is manually annotated, we sample one-tenth of these annotated frames as our test split, resulting in a total of 1,922 test images. For the AGD20K~\cite{luo2022learning} dataset, we use the test split of the \textit{Seen} partition for zero-shot evaluation, which comprises 1,710 object-affordance pairs.

\subsubsection{Baselines.} For a thorough comparison, we evaluate our method against several representative baselines, including open-vocabulary segmentation methods such as VLPart~\cite{sun2023going}, OVSeg~\cite{liang2023open}, and SAN~\cite{xu2023side}; and powerful open-source MLLMs such as LISA~\cite{lai2024lisa}, SAM4MLLM~\cite{chen2024sam4mllm}, AffordanceLLM~\cite{qian2024affordancellm}, Qwen2.5-VL~\cite{qwen25vl}, InternVL3~\cite{internvl3},  Seg-Zero~\cite{Liu2025SegZeroRG}, and Vision Reasoner~\cite{Liu2025VisionReasonerUV} to compare their affordance reasoning capabilities with \textbf{Affordance-R1}.

% \subsubsection{} 
% % gIoU reflects the average of all per-image Intersection-over-Unions (IoUs), while cIoU is defined by the cumulative intersection over the cumulative union. The P@50 metric considers a mask to be a true positive when the IoU ratio exceeds 0.5, and P@50:95 calculates across
% % a range of IoU thresholds from 0.50 to 0.95 with increments of 0.05, then averages across all the thresholds. For the two metric types, IoU and Precision, the latter metric only counts those results greater than a threshold, hence it can pose more challenges to the model and fairly evaluate the results with a high recall rate.

\subsubsection{Evaluation Metrics and Implementation Details.} Following Instruct-Part, we use standard metrics gIoU, cIoU, Precision@50 (P@50), and Precision@50:95 (P@50:95). We employ Qwen2.5-VL-7B~\cite{qwen25vl} and SAM2-Large~\cite{ravi2024sam} as our default configuration. \textbf{Affordance-R1} is trained on a 4×A100 GPU server using the DeepSpeed library. During training, we use a total batch size of 8 with a sampling number of 8 per training step. The initial learning rate is set to 1e-6, the weight decay is 0.01, and the KL loss coefficient is set to 5e-3. The entire training process takes approximately 7 hours.

\subsection{Quantitative Analysis}

We conducted extensive experiments to comprehensively evaluate the affordance reasoning ability of \textbf{Affordance-R1}, including both in-domain and OOD datasets.  
\begin{table}[t]
  \centering
    \resizebox{\linewidth}{!}{
      \begin{tabular}{l|ccccc}
        \toprule
         Model  &Reasoning &gIoU$\uparrow$  & cIoU$\uparrow$ & $P_{50-95}$$\uparrow$ & $P_{50}$$\uparrow$  \\
        \midrule
        LISA-7B  & \usym{2613}&41.90 & 41.23 & 19.33 & 39.65 \\ 
        SAM4MLLM  & \usym{2613} &12.40 & 8.41 & 0.05 & 4.12  \\ 
        AffordanceLLM  & \usym{2613} & 43.11 & 38.97 & 22.36 & 41.56  \\
        Qwen2.5VL-7B  & \usym{2613}  &33.21 & 29.83 & 10.45 & 25.17  \\
        InternVL3-7B & \usym{2613} & 30.46 & 28.73 & 9.94 & 18.67  \\
        Seg-Zero  & \checkmark  &44.26 & 39.30 & 16.53 & 39.93    \\ 
        Vision Reasoner  & \checkmark  &44.00 & 39.71 & 16.10 & 39.04  \\
        \textbf{Affordance-R1}(Ours)  & \checkmark  &\textbf{49.85} & \textbf{42.24} & \textbf{34.08} & \textbf{53.35}  \\
        \bottomrule
      \end{tabular}
  }
  
  \caption{MLLM based zero-shot affordance reasoning comparison results on UMD dataset.}
\label{umd_results}
\end{table}

\subsubsection{Results on \textbf{ReasonAff}.} As presented in Table \ref{main_result}, \textbf{Affordance-R1} establishes a new SOTA on our \textbf{ReasonAff} benchmark, consistently outperforming all baseline methods across every evaluation metric. The performance gains are particularly pronounced on the high-precision metrics, P@50 and P@50:95, underscoring the high quality and accuracy. We show some qualitative comparison results of affordance reasoning in Figure \ref{vis_reasonaff}. More results can be seen in Appendix.

We attribute this superior performance directly to our novel framework. Unlike conventional methods that rely on supervised fine-tuning, \textbf{Affordance-R1} leverages GRPO~\cite{shao2024deepseekmath} to unlock the MLLM's intrinsic reasoning capabilities. This approach is uniquely suited for the challenges posed by \textbf{ReasonAff}, which demands deep reasoning over implicit, complex, and real-world contextual instructions. The core of our success lies in the meticulously designed affordance reward function. Specifically, the Format Reward, which encourages a thinking and rethinking process, compels the model to build a coherent reasoning chain and self-correct before committing to an answer. This iterative refinement process, guided by the Perception and Affordance Recognition rewards, allows \textbf{Affordance-R1} to deconstruct complex problems and accurately ground abstract instructions to visual evidence, a capability where other baselines fall short.

\subsubsection{Results on Out-of-Domain Datasets.} To assess the generalization power of \textbf{Affordance-R1}, we performed a zero-shot evaluation on the AGD20K~\cite{luo2022learning} and UMD~\cite{UAD} datasets. The results, summarized in Table \ref{agd_results} and Table \ref{umd_results}, reveal that \textbf{Affordance-R1} maintains its significant performance edge, demonstrating exceptional generalization to unseen object types and visual domains. This strong generalization is a direct outcome of our methodology. By forgoing traditional SFT in favor of GRPO~\cite{shao2024deepseekmath}, \textbf{Affordance-R1} learns a robust and generalizable policy for affordance reasoning, rather than merely memorizing patterns from the training data. The reinforcement learning process, guided by our comprehensive reward signals, teaches the model the fundamental principles of identifying functional regions based on reasoning. Consequently, this learned policy is less sensitive to domain-specific visual features and translates effectively to novel scenarios presented in OOD datasets. In contrast, competing models show a more significant performance drop, indicating a degree of overfitting to their training distributions and a weaker grasp of the underlying affordance concepts. This confirms that \textbf{Affordance-R1} learns a more fundamental and transferable understanding of object affordance.
 
% \begin{table}[t]
%   \centering
% \resizebox{\linewidth}{!}{

%       \begin{tabular}{l|cccccc}
%         \toprule
%          Model & Type & LLM &Reasoning  & KLD$\downarrow$ & SIM$\uparrow$ & NSS$\uparrow$  \\
%         \midrule
%         InteractionHotspots & Supervised &\usym{2613} & \usym{2613} &1.994& 0.237& 0.577 \\
%         Cross-View-AG  & Supervised &\usym{2613} & \usym{2613}  &1.787 &0.285& 0.829\\
%         AffCorrs & Supervised &\usym{2613} & \usym{2613} &1.618 &0.348 &1.021 \\
%         LOCATE   & Supervised &\usym{2613} & \usym{2613} &1.405 &0.372 &1.157 \\
%         AffordanceLLM & Supervised &\checkmark & \usym{2613} &1.463 &0.377 &1.070 \\
%         \midrule
%         LISA-7B &Zero Shot &\checkmark & \usym{2613}& 13.68 & 0.16 & 0.46 \\ 
%         SAM4MLLM &Zero Shot &\checkmark & \usym{2613} & 9.51 & 0.27 & 0.52  \\ 
%         Qwen2.5VL-7B &Zero Shot &\checkmark & \usym{2613}  &9.81 &0.26 &0.77\\
%         InternVL3-7B &Zero Shot &\checkmark & \usym{2613}   &10.09 &0.25 &0.61\\
%         Seg-Zero &Zero Shot &\checkmark  & \checkmark  & \textbf{9.02} & 0.35 & 0.94  \\
%         Vision Reasoner &Zero Shot &\checkmark & \checkmark  & 8.90 & 0.35 & 0.95  \\
%         \textbf{Affordance-R1}(Ours) &Zero Shot &\checkmark & \checkmark  &9.73 & \textbf{0.36} & \textbf{0.98} \\
%         \bottomrule

%       \end{tabular}}
    
%   \caption{Comparison Results on AGD20K Dataset.}
% \label{agd_results}
% \end{table}

\begin{table}[t]
  \centering
    \resizebox{\linewidth}{!}{
      \begin{tabular}{l|ccccccccc}
        \toprule
         Model  &Reasoning &gIoU$\uparrow$  & cIoU$\uparrow$ & $P_{50-95}$$\uparrow$ & $P_{50}$$\uparrow$ & KLD$\downarrow$ & SIM$\uparrow$ & NSS$\uparrow$  \\
        \midrule
            LISA-7B  & \usym{2613}& 13.18 & 11.96 & 1.45 & 5.31 & 13.68 & 0.16 & 0.46 \\ 
        SAM4MLLM  & \usym{2613} & 15.27 & 13.22 & 2.40 & 6.95 & 9.51 & 0.27 & 0.52  \\ 
        % AffordancLLM  & \usym{2613} &  & 0 & 0 & 0 &0 &0 &0 \\
        Qwen2.5VL-7B  & \usym{2613} &20.28 & 16.35 & 5.61 & 15.49  &9.81 &0.26 &0.77\\
        InternVL3-7B  & \usym{2613} & 18.18 & 14.63 & 3.79 & 13.37  &10.09 &0.25 &0.61\\
        Seg-Zero  & \checkmark  &26.99 & 22.01 & 6.52 & 17.82 & 9.02 & 0.35 & 0.94  \\
        Vision Reasoner  & \checkmark  &26.98 & 21.98 & 6.31 & 17.31 & \textbf{8.90} & 0.35 & 0.95  \\
        \textbf{Affordance-R1}(Ours)  & \checkmark &\textbf{31.78} & \textbf{27.85} & \textbf{7.99}& \textbf{20.49}  &9.73 & \textbf{0.36} &\textbf{0 .98}\\
        \bottomrule

      \end{tabular}}

  \caption{MLLM based zero shot affordance reasoning comparison results on AGD20K dataset.}
\label{agd_results}
\end{table}

\subsubsection{Visualization Results on Web Image.}
\begin{figure}
    \centering
    \includegraphics[width=0.9\linewidth]{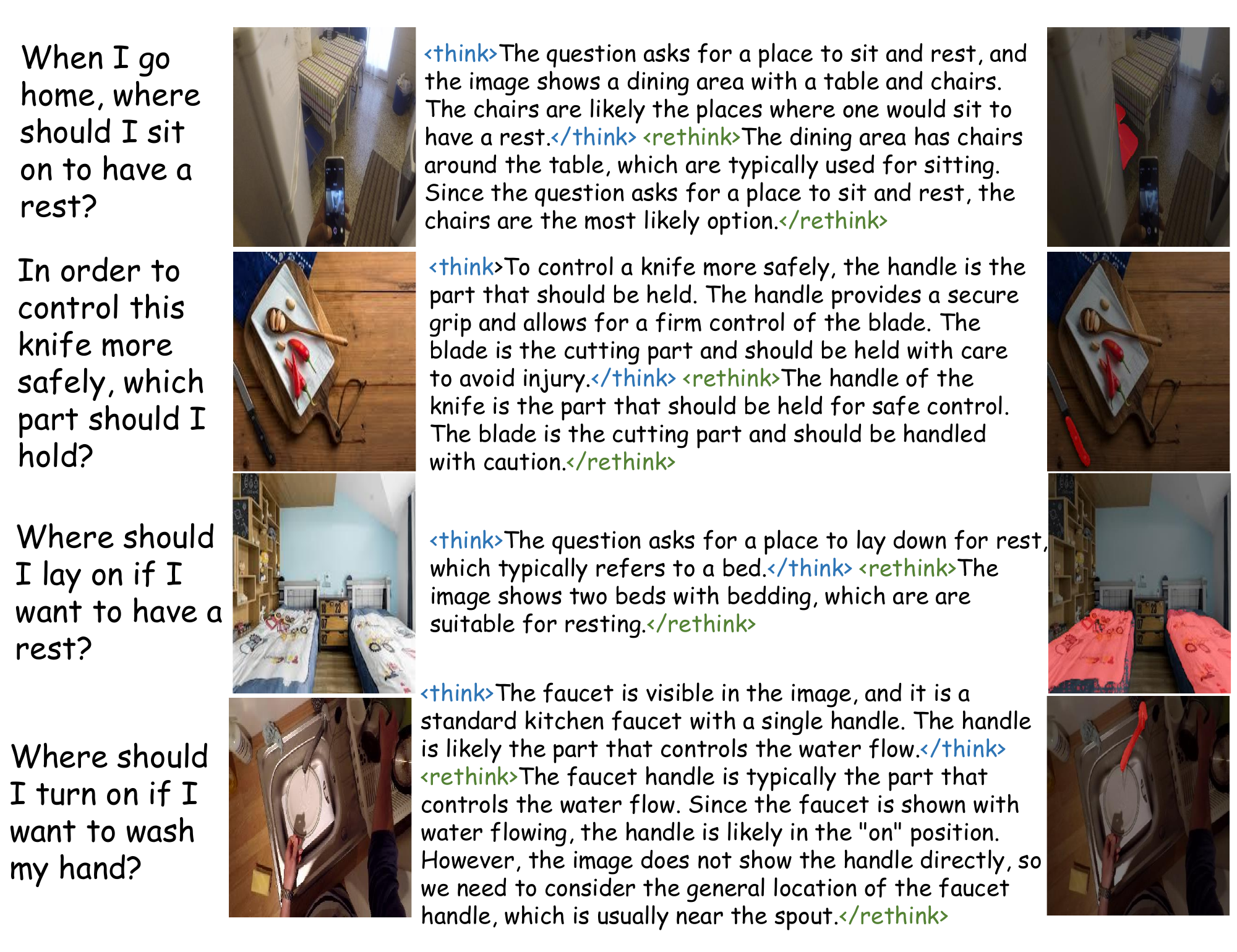}
    \caption{\textbf{Visualization on Web Image.} \textbf{Affordance-R1} can understand complex scenarios and shows well generalization.}
    \label{web}
\end{figure}

To evaluate the generalization ability of \textbf{Affordance-R1}, we collect some kitchen and household scene pictures from the EPIC-KITCHENS dataset~\cite{damen2018scaling} and the internet. As can be seen in Figure \ref{web}, \textbf{Affordance-R1} can still maintain strong affordance reasoning ability and effectively handle complex scenarios. More results can be seen in Appendix.

\subsection{Ablation Study Results}
We conduct various ablation studies to assess the impact of different components on our model \textbf{Affordance-R1}'s performance, including the proposed rethinking reward, the affordance recognition reward, and the Box-Num reward. 

\begin{table}
\centering

\resizebox{\linewidth}{!}{

\begin{tabular}{ccc|cccc}
\toprule
 \textbf{Rethinking}& \textbf{Recognition} & \textbf{Box-Num } &gIoU$\uparrow$  & cIoU$\uparrow$ & $P_{50}$ $\uparrow$ & $P_{50-95}$$\uparrow$   \\ 
\midrule
\usym{2613}  & \usym{2613} & \usym{2613} &60.58 & 51.94 & 66.89 & 45.55 \\
 \checkmark & \usym{2613} &\usym{2613}  &63.04 & 56.33 & 67.02 & 51.55 \\
 \checkmark& \checkmark & \usym{2613}  &65.25 & 61.22 & 68.33 & 50.07 \\
 \checkmark & \checkmark &\checkmark   &\textbf{67.41} & \textbf{62.72} & \textbf{74.50} & \textbf{55.22} \\ 

 \bottomrule
\end{tabular}}

\caption{\textbf{Ablation Study.} We investigate the improvement of Rethinking Reward and Affordance Reward on the model performance based on the baseline. 
}
\label{ablation}

\end{table}

\subsubsection{Rethinking Reward.} As ancient wisdom states: \textit{``Think twice before you act''}. The results Table \ref{ablation} demonstrate that the introduction of the rethinking reward can force the model to reconsider and re-examine the question and image, making it think twice before giving final answers, resulting in an improvement over the baseline.

\subsubsection{Affordance Recognition Reward.} As the saying goes, \textit{``to know what it is and to know why it is''}, affordance reasoning not only requires the model to know where the affordance area is but also the type of affordance this object affords. Table \ref{ablation} presents the performance comparison with and without the affordance recognition reward. The model achieves better results when trained using the affordance recognition reward, which means the affordance recognition reward can help the model understand the concept of affordance and general affordance knowledge.

\subsubsection{Box-Num Reward.} As can be seen in Table \ref{ablation}, we conducted ablation experiments to study the influence of the box-num reward. We found that without this reward function, the model would tend to output a single affordance reasoning answer and ignore other possibilities, resulting in performance degradation.

% \textbf{Model Size}

% \begin{table*}[t]
%   \centering
%   \scalebox{1.0}{
%       \begin{tabular}{l|ccccccccc}
%         \toprule
%          Model & LLM &Reasoning &gIoU  & cIoU & $P_{50-95}$ & $P_{50}$ & KLD$\downarrow$ & SIM$\uparrow$ & NSS$\uparrow$  \\
%         \midrule
%         LISA-7B & \usym{2613} & \usym{2613}& 13.18 & 11.96 & 1.45 & 5.31 & 13.68 & 0.16 & 0.46 \\ 
%         SAM4MLLM & \usym{2613} & \usym{2613} & 15.27 & 13.22 & 2.40 & 6.95 & 9.51 & 0.27 & 0.52  \\ 
%         AffordancLLM & \checkmark & \usym{2613} & 0 & 0 & 0 & 0 &0 &0 &0 \\
%         Qwen2.5VL-7B & \checkmark & \usym{2613} &20.28 & 16.35 & 5.61 & 15.49  &9.81 &0.26 &0.77\\
%         InternVL3-7B & \checkmark & \usym{2613} & 18.18 & 14.63 & 3.79 & 13.37  &10.09 &0.25 &0.61\\
%         Seg-Zero & \checkmark  & \checkmark  &26.99 & 22.01 & 6.52 & 17.82 & 9.02 & 0.35 & 0.94  \\
%         Vision Reasoner & \checkmark & \checkmark  &26.98 & 21.98 & 6.31 & 17.31 & 8.90 & 0.35 & 0.95  \\
%         \textbf{Affordance-R1}(Ours) & \checkmark & \checkmark &0 & 0 & 0 & 0 &0 & 0 &0 \\
%         \bottomrule

%       \end{tabular}}

%   \caption{MLLM based zero shot affordance reasoning comparison results on AGD20K dataset.}

% \end{table*}

%% file: sec/conclusion.tex
\section{Conclusion and Future Work}

In this paper, we introduce the first affordance-centric reasoning model \textbf{Affordance-R1} and a high-quality affordance-centric reasoning dataset \textbf{ReasonAff}, which can be integrated into the instruction-tuning training process of multimodal large language models. With the help of the proposed sophisticated affordance reasoning reward function, we adopt pure reinforcement learning, specifically GRPO, to fine-tune the MLLM without supervised fine-tuning (SFT). \textbf{Affordance-R1} advances affordance reasoning by integrating LLM capabilities, enhancing the model's ability to handle complex and real-world contexts. It not only achieves state-of-the-art performance on \textbf{ReasonAff} but also shows superior generalization on out-of-domain datasets. For future work, we will explore how to utilize the excellent affordance reasoning abilities of \textbf{Affordance-R1} to construct an automatic data engine pipeline for affordance reasoning, thereby advancing the scaling law of embodied perception.

%% file: sec/sup.tex
\clearpage
\appendix

\section*{Supplementary Material }

This document contains supplementary materials for our main paper. We provide further technical details and more qualitative examples to complement the findings presented in the main text. We hope this supplementary information will help readers better understand our approach and results.

The remainder of this supplementary material is organized as follows. In Section A, we provide the hardware specifications used in our experiments. In Section B, we list the hyperparameters employed. Section C presents the detailed prompts for training and inference. Section D shows more details about the dataset. Section E gives a detailed picture of the future affordance data engine. Section F presents more visualizations. 

\section{ Computational Resources}
\label{Computational Resources}

To ensure reproducibility, we provide detailed information on the computational resources used in our experiments. For all experiments, including training and inference, we used 4 NVIDIA RTX A800 GPUs. The base model we used is Qwen-2-VL-7B, consuming approximately 78GB of memory during operation.

\section{Hyperparameters}

In Table \ref{hyper}, we present the hyperparameters used in our experiments. 
\begin{table}[ht]

\centering
\begin{tabular}{@{}l|c@{}}
\toprule

Hyperparameter  &  Value  \\
\midrule
Batch Size  & 8   \\
Experience  &  16  \\
Tempture    & 1  \\
KL  & 0.05   \\
epoch   & 5  \\
RL Steps &750  \\
Optimizer & AdamW   \\
Learning Rate & 1.e-4    \\
seed & 42\\
max pixels& 12845056\\
min pixels& 3136\\
max response length& 2048\\
weight decay &  1.0e-2\\
max promptlength& 1300\\
\bottomrule
\end{tabular}

\caption{Details of  Hyperparameters
}
\label{hyper}

\end{table}

\section{ Prompts }

We have carefully designed prompts to enable MLLM to build datasets and complete affordance reasoning.

\subsection{Prompts for Data Construction}

\begin{figure}[t]
    \centering
    \includegraphics[width=0.8\linewidth]{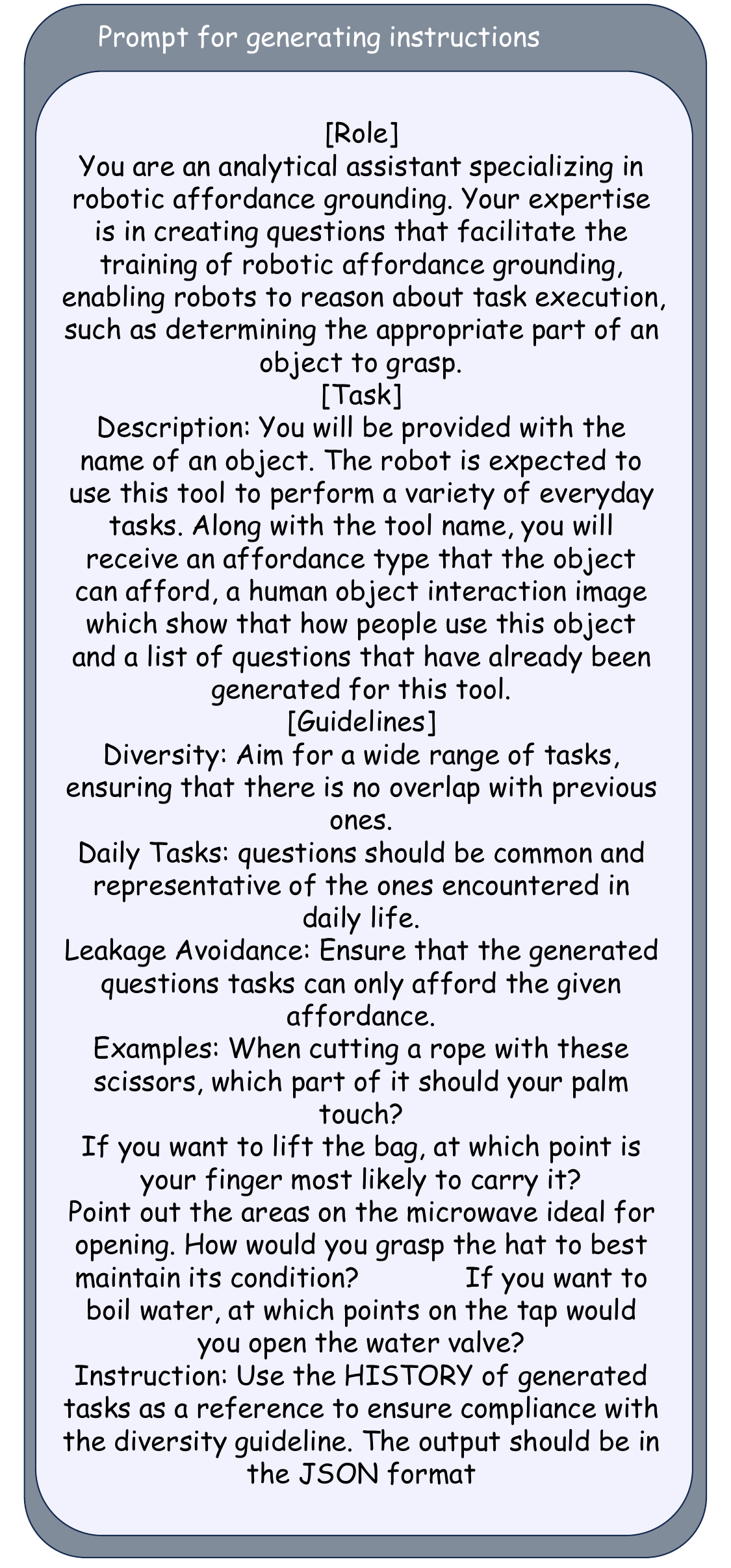}
    \caption{Full prompt for generating reasoning instruction. }
    \label{prompt_instruction}
\end{figure}

\paragraph{Prompt for generating reasoning-based instructions.} We hope that the model can provide complex instructions based on different contextual scenarios. We collect Human-Object-Interaction images to prompt the GPT-4o to relieve hallucination problems. The details of our prompt are shown in Figure \ref{prompt_instruction}.

\begin{figure}[t]
    \centering
    \includegraphics[width=0.7\linewidth]{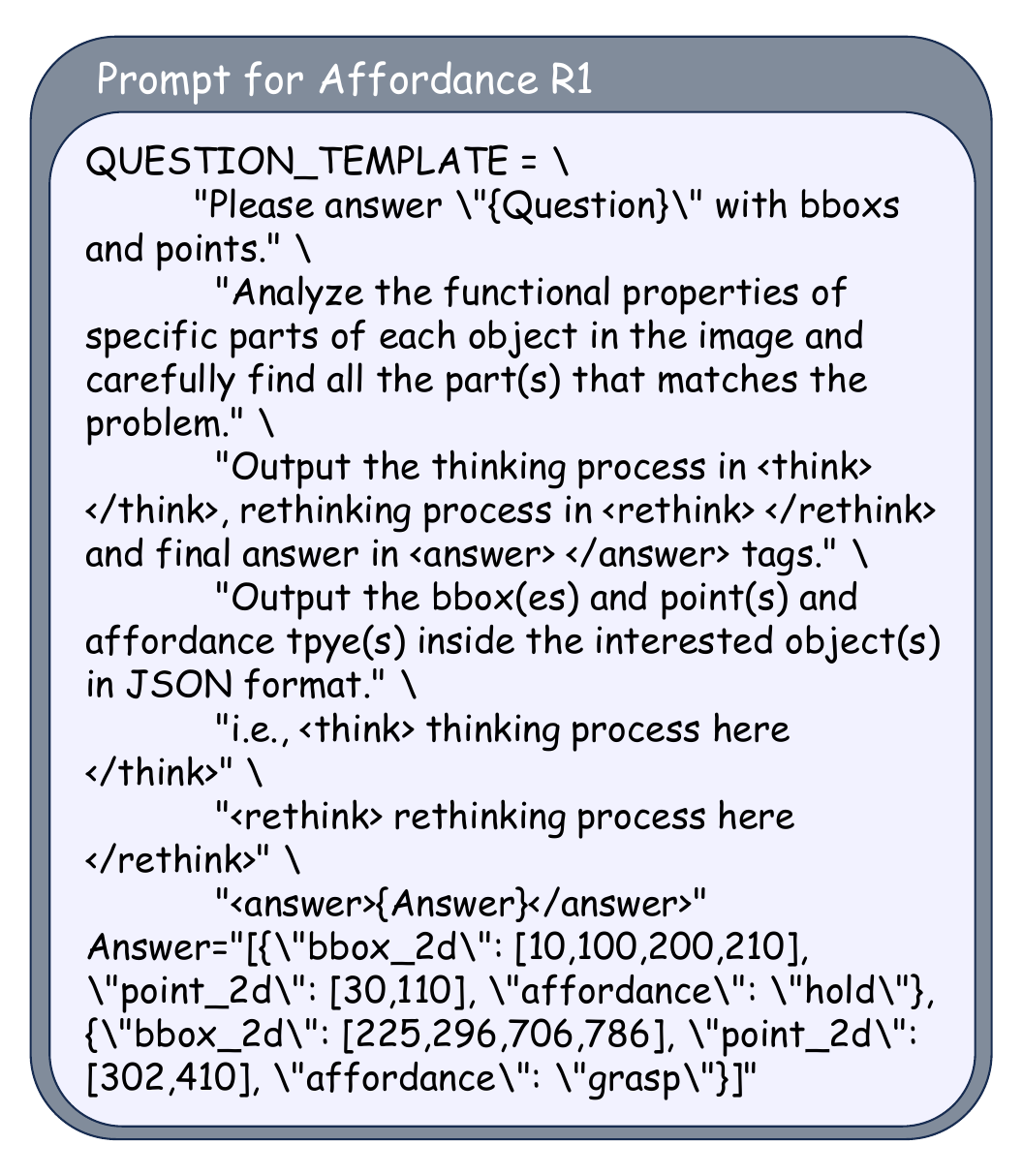}
    \caption{Full prompt for affordance training and inference.}
    \label{prompt_instructionr1}
\end{figure}

\paragraph{Prompt for affordance reasoning.} In order to better stimulate the reasoning ability of the large model during training, we have carefully designed prompts to guide the model and provide a specific answer example to help the model understand the task. The details of our prompt are shown in Figure \ref{prompt_instructionr1}.

\section{ Datasets }

In this section, we give more details about the proposed dataset, ReasonAff.

\subsection{ Dataset Details}
\begin{figure}
    \centering
    \includegraphics[width=0.9\linewidth]{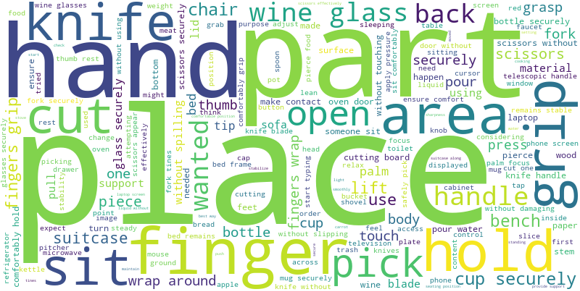}
    \caption{Word could of ReasonAff}
    \label{fig:word}
\end{figure}

\begin{figure}
    \centering
    \includegraphics[width=0.9\linewidth]{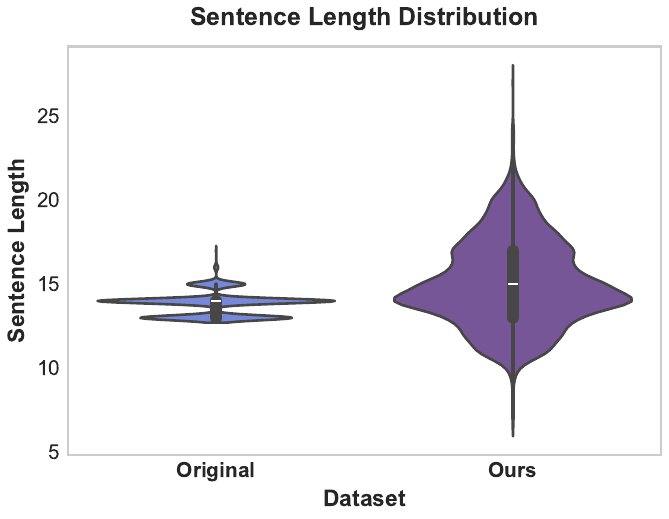}
    \caption{Comparison of violin plots of instruction length between raw data and Reasonaff}
    \label{fig:violin}
\end{figure}

In order to more intuitively demonstrate the diversity and superiority of our proposed dataset, ReasonAff, we calculated the distribution of word frequency and instruction length for the data instruction. Figure \ref{fig:word} shows the word cloud of the generated instructions, and Figure \ref{fig:violin} shows the comparison of violin plots of instruction length between raw data and Reasonaff. The superiority of our data is reflected in the more diverse reasoning-based affordance instructions, which take into account rich scene contexts and make MLLM more suitable for real-world scenarios

\section{Future work:Affordance Data Engine }

\begin{figure}
    \centering
    \includegraphics[width=0.9\linewidth]{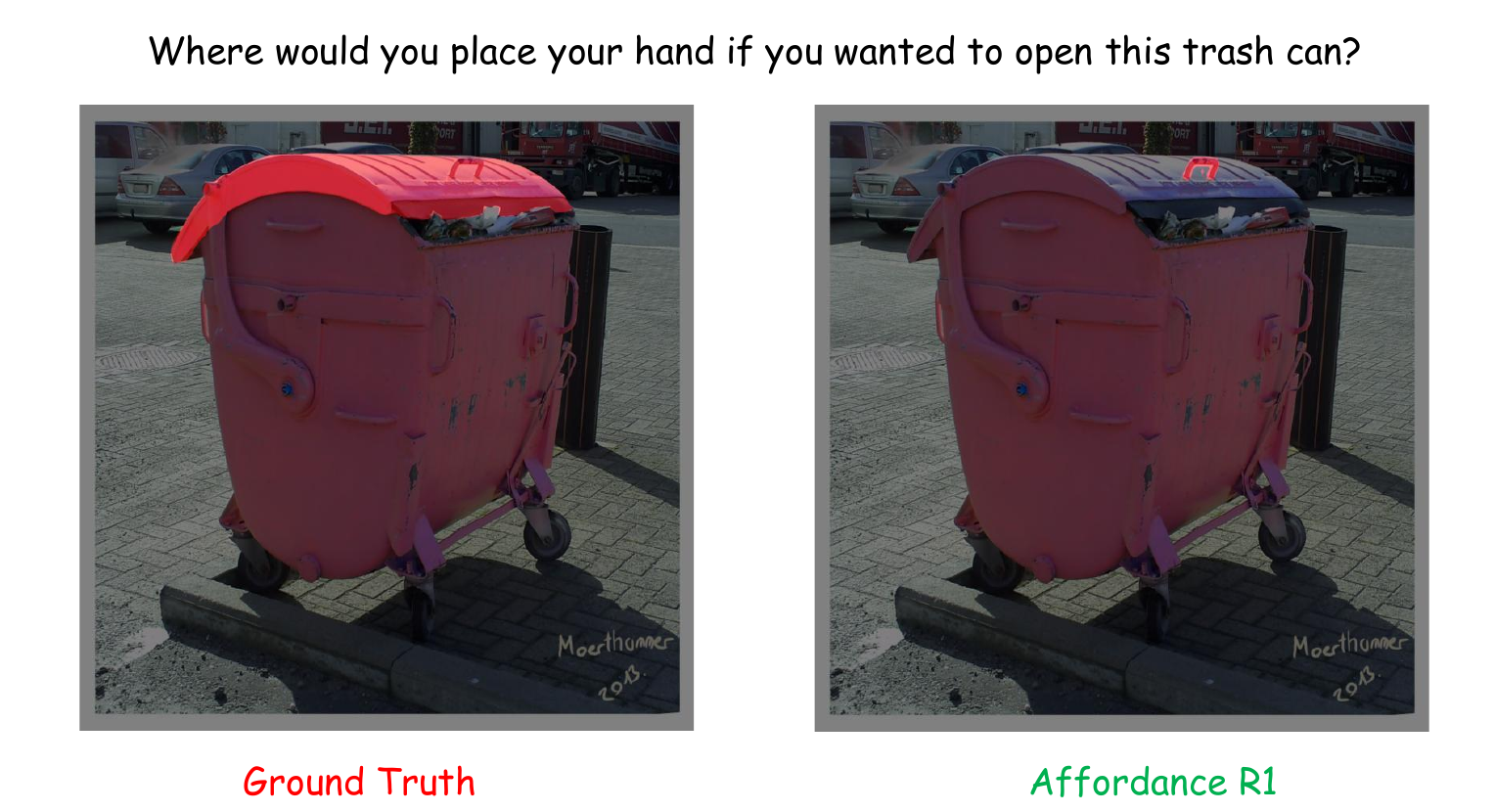}
    \caption{Ground truth is coarser in some cases, and Affordance R1 may predict a finer-grained affordance mask.}
    \label{fig:data_engine}
\end{figure}

\begin{figure}
    \centering
    \includegraphics[width=\linewidth]{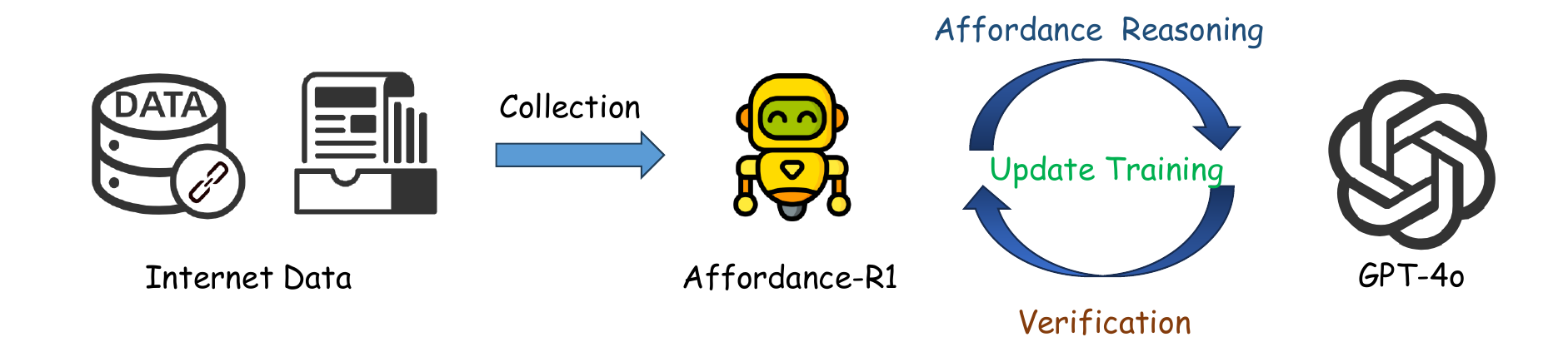}
    \caption{Model in the loop scaling up for affordance reasoning.}
    \label{fig:modelinloop}
\end{figure}

Although ReasonAff provides a good source of affordance reasoning data, its scale and granularity are still insufficient. We built ReasonAff based on InstructParts, but it focuses more on part-level affordance. During the experiment, we found that its annotation of partial affordance data was coarser. As can be seen in Figure \ref{fig:data_engine}, in this case, the instruction is \textit{ Where would you place your hand if you wanted to open this trash can? }, the ground truth highlights the lid of the trash can, and our Affordance-R1 predicts more accurately. This enables us to utilize the excellent affordance reasoning ability of Affordance-R1 to make it an affordance data engine, which can predict the finer-grained affordance mask of the given, in-the-wild image. We hope to adopt the Affordance as the affordance mask generator, and an advancing VLM (e.g, GPT-4o) as the judge, inspired by the \textit{MLLM-as-a-Judge} strategy. The judge VLM may not own such wonderful affordance grounding ability, but it is reasonable to adopt it as the \textbf{verifier} to filter the output of Affordance-R1. We would like to further explore this by utilizing the strategy of \textbf{Model-in-the-Loop} to construct a large-scale affordance reasoning dataset, which is suitable for MLLM to perform affordance reasoning instruction-tuning, to scale up the scaling law in embodied perception. As shown in Figure \ref{fig:modelinloop}, we would use the filter data to further train the model, improving the affordance reasoning ability. The data may come from the internet or even data generated by generative models. 

\section{ More Visualization}

In this section, we provide more visualization of Affordance-R1, the results are as follows

 \begin{figure*}[t]
    \centering
    \includegraphics[width=\linewidth]{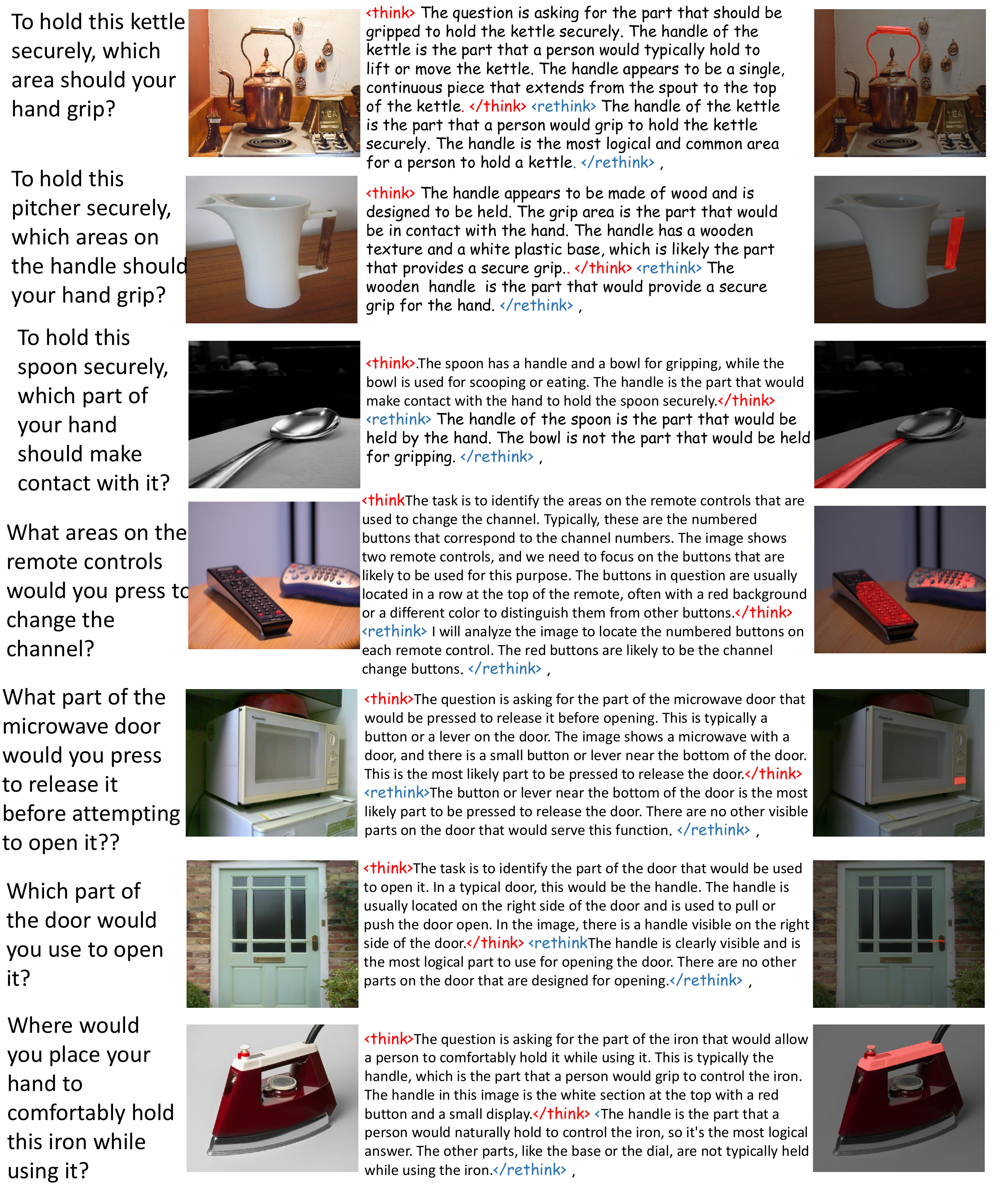}s
    \caption{Visualization on ReasonAff. \textbf{Affordance-R1} can understand complex scenarios and shows well generalization. }
\end{figure*}

 \begin{figure*}[t]
    \centering
    \includegraphics[width=\linewidth]{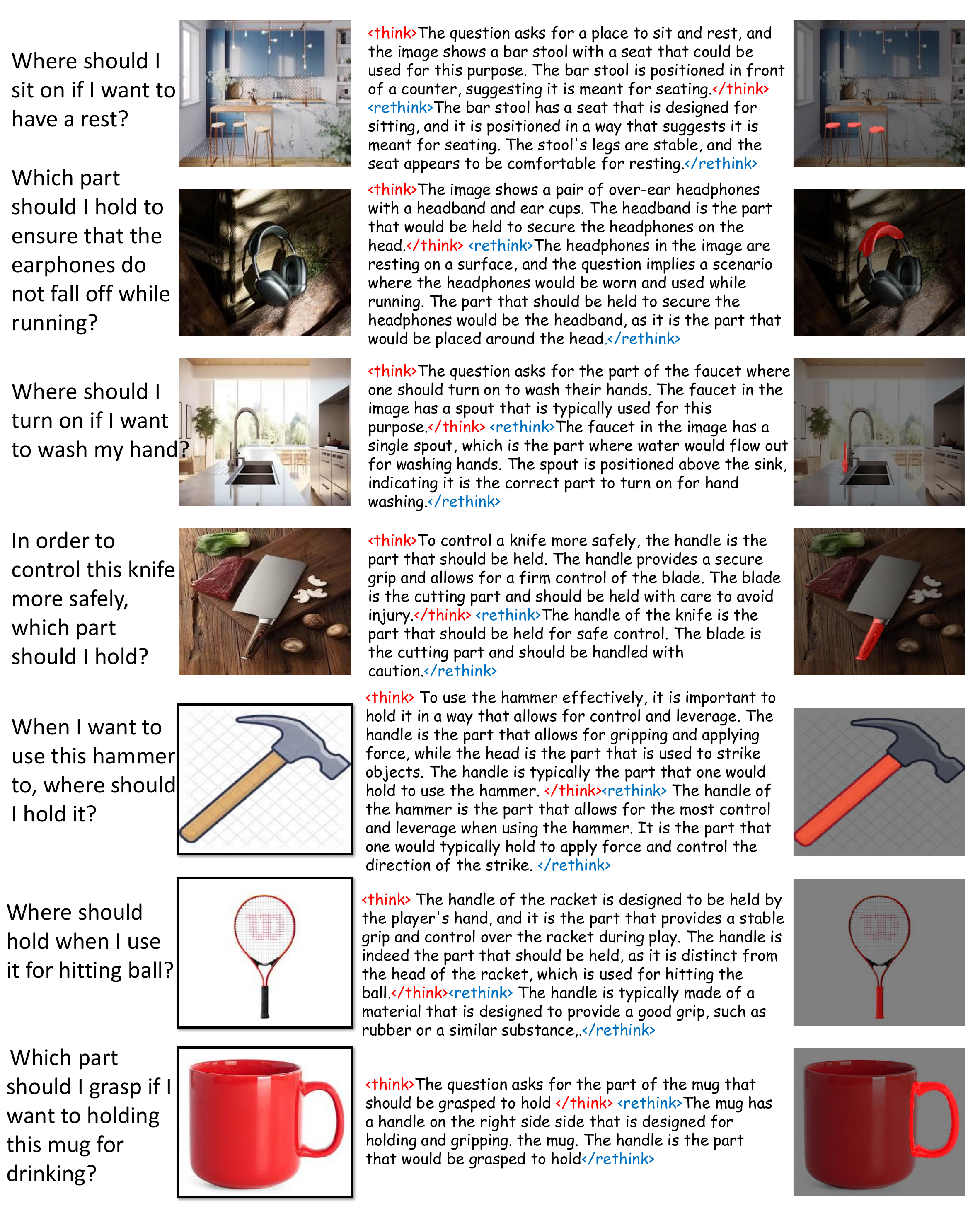}s
    \caption{Visualization on Web Image. \textbf{Affordance-R1} can understand complex scenarios and shows well generalization. }
\end{figure*}